\documentclass[runningheads]{llncs}
\usepackage{longtable}

\usepackage{textcomp}
\usepackage{pgfplots}
\usepgfplotslibrary{external}
\usepackage{cite}
\usepackage{array}
\usepackage{booktabs}
\usepackage{tikz}
\usetikzlibrary{shapes.geometric, arrows}
\usepackage{smartdiagram}
\usesmartdiagramlibrary{additions}
\usepackage{hhline}
\usepackage{multirow}
\usepackage{amsmath}
\usepackage[nomessages]{fp} %Math
\usepackage{dblfloatfix} %Lets putting * tables and figures at bottom

\begin{document}
% \frontmatter

\title{User Localization Based on Call Detail Record}

\author{
Buddhi Ayesha\inst{1}\and
Bhagya Jeewanthi\inst{2}\and
Charith Chitraranjan\inst{1}\and
Amal Shehan Perera\inst{1}\and
Amal S. Kumarage\inst{2}
}
\authorrunning{B. Ayesha et al.}
\institute{Department of Computer Science \& Engineering, University of Moratuwa
\and
Department of Transport \& Logistics Management, University of Moratuwa\\
\email{buddhiayesha.13@cse.mrt.ac.lk}
}

% \author{Anonymous Authors}

\maketitle

\begin{abstract}
Understanding human mobility is essential for many fields, including transportation planning. Currently, surveys are the primary source for such analysis. However, in the recent past, many researchers have focused on Call Detail Records (CDR) for identifying travel patterns. CDRs have shown correlation to human mobility behavior. However, one of the main issues in using CDR data is that it is difficult to identify the precise location of the user due to the low spacial resolution of the data and other artifacts such as the load sharing effect. Existing approaches have certain limitations. Previous studies using CDRs do not consider the transmit power of cell towers when localizing the users and use an oversimplified approach to identify load sharing effects. Furthermore, they consider the entire population of users as one group neglecting the differences in mobility patterns of different segments of users. This research introduces a novel methodology to user position localization from CDRs through improved detection of load sharing effects, by taking the transmit power into account, and segmenting the users into distinct groups for the purpose of learning any parameters of the model. Moreover, this research uses several methods to address the existing limitations and validate the generated results using nearly 4 billion CDR data points with travel survey data and voluntarily collected mobile data.


\end{abstract}

%Variables 
\newcommand{\ranf}{$RF{\scriptscriptstyle}$}
\newcommand{\ann}{$ANN{\scriptscriptstyle}$}
\newcommand{\svm}{$SVM_{\scriptscriptstyle WC}$}

%	~\cite{}

\section{Introduction}
%Begin: Introduction %%%%%%%%%%%%%%%%%%%%%%%%%%%%%%%%%%%%%%%%%%%%%%%%%%%%%%%%%%%%%%%%%%%%%%%%%%%%%%%%%%%%%%%%%%%%%%%%%%%%%%%%%%%%%%%%%%%%%%%%%%%%%%%%%%%%%%%%%%%%%%

Human mobility patterns are critical sources of information for designing, analyzing and enhancing transport planning activities. These transportation initiatives require mobility rich data to become a continuous process. Currently, manually carried out roadside surveys or household surveys are done once every few years provide input data for such initiatives. Such data collection while being expensive, takes time and is often outdated by the time it is available for analysis. Such drawbacks of the existing methodologies create the necessity for more advanced data extraction techniques to analyze human motion.

The recent adoption of ubiquitous computing technologies, including mobile devices by very large portions of the world population has enabled the capturing of large-scale spatio-temporal data about human motion~\cite{vieira2010querying}. Large-scale penetration of mobile phones serves as a dynamic source to derive insights on human mobility due to the generation of big data. Mobile Network Big Data makes it possible to locate mobile devices in time and space with a certain accuracy using the mobile network infrastructure that includes location information of the mobile device~\cite{tiru2014overview}. In the present scenario, Call Detail Records (CDR) data, which is collected by mobile phone carriers for billing purposes is the most common type of MNBD used in a variety of transportation studies. CDRs generate the time-stamped locations of the users. All the phone numbers are replaced by a computer generated unique identifier by the operator to preserve the anonymity and the privacy of the users. A single CDR contains a random ID number of the phone, device and phone number, exact time and date, call duration, location in terms of latitude and longitude of the cell tower that provide the network signal. When a person makes a call, all this information gets recorded as an array of data. These arrays of data provide information on the mobility of the users based on the places they have travelled.

\begin{figure}[!hb]	
	\centering
\includegraphics[width=0.4\textwidth]{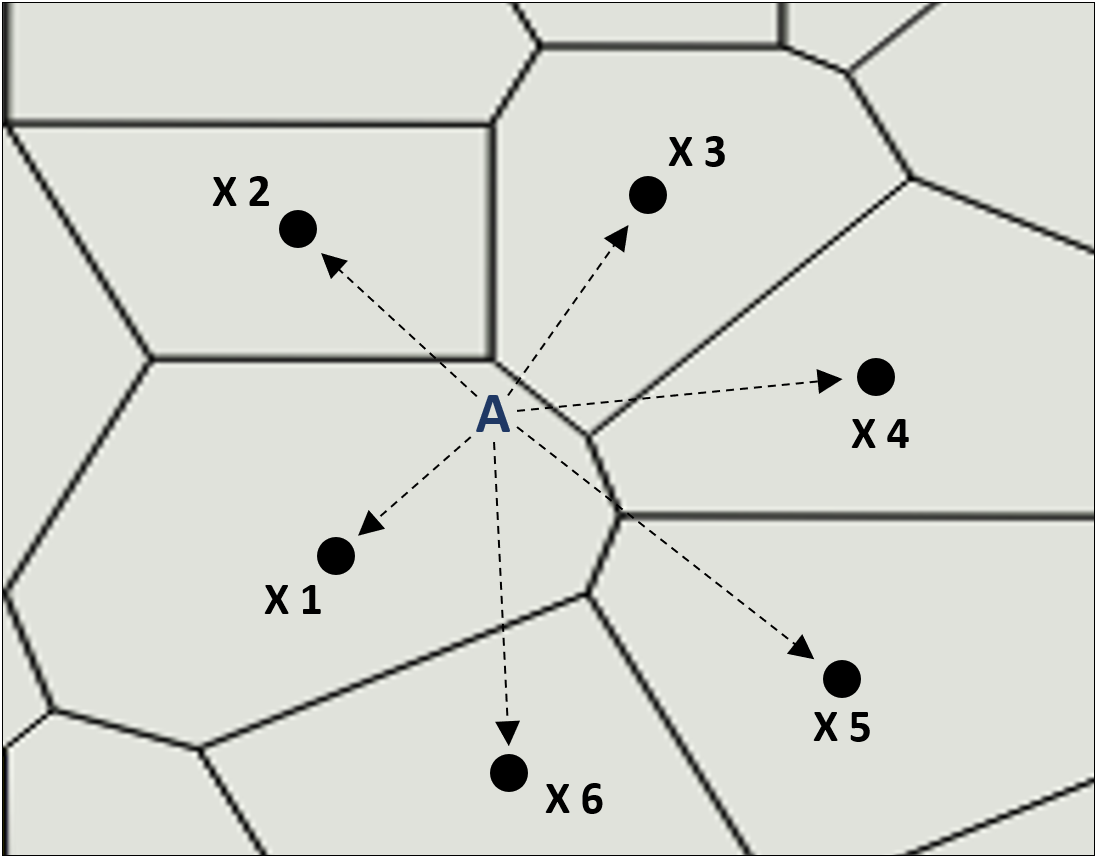}
	\caption{Load sharing effect}	
	\label{fig:load_sharing_effect}
\end{figure}

Though CDR has the potential to provide information on human mobility insights at a wider scope, there are certain limitations embedded within CDR data that makes the accuracy of the derived information to be low. One main limitation of the CDR data is that for a given call, there may be frequent changes in the serving tower with no actual displacement of the user, which is referred to as the load sharing effect ~\cite{iqbal2014development}. This is mainly because the operator often balances call traffic among adjacent towers by allocating a new call to the tower that is handling a lower call volume at that moment. As an example, ~\figurename~\ref{fig:load_sharing_effect} indicates the general demonstration of cell towers. User has the possibility of connecting to any tower depicted. But the tower is allocated by the operator based on the total number of calls each one can handle. With that, user might get connected to X1 at one time and next time to X2, even though the A's location does not change.

Due to this process, the towers providing the signal varies without actual displacement of the user. This has a large effect on the transportation aspect since the trip counts between origins and destinations may exceed actual values, behavioral structures may incorrectly be identified and etc. Therefore, it is mandatory to fix the existing issues in order to use these real-time data for transportation forecasting processes.

Other than the load sharing effect, one other main issue is the localization error, which means the user's location is identified as anywhere within the area of the connected cell tower. For an example, user A's (~\figurename~\ref{fig:load_sharing_effect}) location is identified as anywhere within the coverage area of X1, X2 or X6. Existing studies do not take into account, the load sharing effect or the transmit power of the connected cell tower when localizing the user. The study propose an improved method to localize the user by incorporating the load sharing effect as well as the transmit power levels of the cell towers. Furthermore, when it is required to fit any model parameters to the data, the first attempt is to identify separate groups of users based on demographic attributes and fit the parameters separately to different groups as opposed to the conventional approach of fitting parameters to the entire population of users as one.  

This study demonstrates a method to reduce the localization error from CDR data. Second section presents an overview of the previous works done and the drawbacks of the existing studies. Section 3 summarizes the methodology of the research conducted, and section 4 validate the results against the independent data sources from the study area. Based on these findings, the study concludes with a discussion of the limitations and applications of CDR data in future studies.
%End: Introduction %%%%%%%%%%%%%%%%%%%%%%%%%%%%%%%%%%%%%%%%%%%%%%%%%%%%%%%%%%%%%%%%%%%%%%%%%%%%%%%%%%%%%%%%%%%%%%%%%%%%%%%%%%%%%%%%%%%%%%%%%%%%%%%%%%%%%%%%%%%%%%%%%%

\section{Background and Related Work}
\label{Background}
%Begin: Background and Related Work %%%%%%%%%%%%%%%%%%%%%%%%%%%%%%%%%%%%%%%%%%%%%%%%%%%%%%%%%%%%%%%%%%%%%%%%%%%%%%%%%%%%%%%%%%%%%%%%%%%%%%%%%%%%%%%%%%%%%%%%%%%%%%%%

Previous studies have used different methodologies to minimize the load sharing effect. Basically there are two techniques that have been used as trajectory smoothing and spatial clustering.

\subsection{Trajectory Smoothing}
In trajectory smoothing, the sequence of CDR within a certain time threshold is taken into consideration and different smoothing or filtering algorithms were taken into account to reduce the “jumps” in the location sequence. Speed based filtering, time weighted smoothing and assigning a single medoid location to records which are close by were different algorithms used in trajectory smoothing to smooth the location sequence.

In speed based filtering a small trajectory of data points within five minutes are inspected and any points which have irrational travel distance or travel speed (more than 120 kmph) are removed from the database ~\cite{wang2013estimating}. In the studies which use time weighted smoothing, the preprocessing is initiated by smoothing the positions independently for each user. In time weighted smoothing, the maximum stay region size is set to d = 300m to approximate the area that might likely be traversed on foot as part of an urban activity. Then the entire region is divided into rectangular cells of size d/3. Then, all the stay points were mapped and iteratively merge the unlabeled cell with the maximum stay-points and its unlabeled neighbors to a new stay-region and labeled the stay regions~\cite{jiang2013review}.

In conclusion, the trajectory smoothing also has its own drawbacks. The user visited locations are removed in the trajectory smoothing method which minimize the possibility to identify exact locations of the users

\subsection{Spatial Clustering}
Spatial clustering is the other main technique that is used to remove the load sharing effect. There the data points were clustered based on the spatial distribution without considering the temporal distribution of CDR. This process allows to consolidate points that may represent the same location but, visited in different days. Agglomerative~\cite{hariharan2004project} and leader clustering~\cite{isaacman2011identifying} are the two main clustering techniques used for this categorization.

Both these methodologies have their own drawbacks. Initially the speed between two consecutive records were calculated using the time difference and the distance parameter. The distance between two consecutive points is the Euclidean distance. Then in speed based filtering of trajectory smoothing, only the data within five minutes were inspected, where the travel speed considered is 120 miles per hour. But the speed should be sensitivity tested rather than using a selected speed.
 
In trajectory smoothing the positions were smoothed based on a weight. The positions were smoothed for all users based on their caller activities.
 
In spatial clustering, a stay-point is identified by a sequence of consecutive cell phone records bounded by both temporal and spatial constraints. The spatial constraint is the roaming distance when a user is staying at a location, which should be related to the accuracy of the device collecting location data. The temporal constraint is the minimum duration spent at a location, which is measured as the temporal difference between the first and the last record in a stay. In most of the studies that used spatial clustering techniques for the load sharing removal, only the latitudinal and longitudinal locations were considered without taking the number of caller activities in each location into consideration. Leader clustering is one of the techniques used in spatial clustering, initially the cell towers are sorted and one with the most counts is taken as the centroid of the first cluster. Then, for each subsequent cell tower, it checks whether the tower falls within a threshold radius of the centroid of any existing cluster. If it does not, the tower becomes the centroid of a new cluster. If it does fall within the threshold radius of an existing cluster, the algorithm adds the tower to the cluster and moves the centroid of the cluster to be the weighted average of the locations of all the cell towers in the cluster. Choosing a particular threshold radius around cell towers helps to equalize for the fact that in urban areas towers might be as dense as 200 meters apart, while in suburban areas, a spacing of 1-3 miles are more common. These threshold values should be sensitivity tested and validated with another data source, to be used in the study
 
Additionally, with respect to spatial clustering only the spatial behavior is taken into account, without considering about the time factor. Specially the sequence of location visits were not considered. Other main limitation in literature is the lack of models to minimize location error using label data.

In summary, There are three main influencing factors for the localization to be stated as load sharing, signal strength of the cell tower and the frequency of connecting to a particular tower. Existing studies has a gap where they have not considered all these three factors together to remove the localization error. Additionally this study combines with user profiling for the localization error reduction in a novel manner.

%End: Background and Related Work %%%%%%%%%%%%%%%%%%%%%%%%%%%%%%%%%%%%%%%%%%%%%%%%%%%%%%%%%%%%%%%%%%%%%%%%%%%%%%%%%%%%%%%%%%%%%%%%%%%%%%%%%%%%%%%%%%%%%%%%%%%%%%%%

\section{Methodology}
\label{Methodology}

\begin{figure*}[!htb]	
	\centering
\includegraphics[width=1\textwidth]{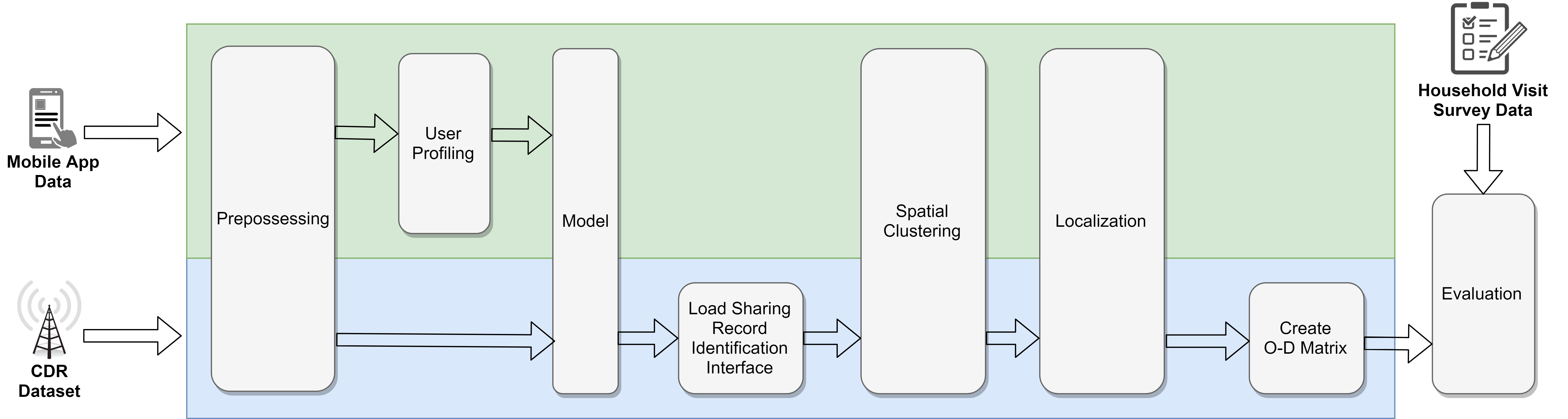}
	\caption{Flow diagram for the overall methodology}	
	\label{fig:Overall methodology}
\end{figure*}

This section describes the overall research methodology that was carried out. Each section below, addresses a component in the overall methodology. An overview of the methodology proposed is illustrated in~\figurename~\ref{fig:Overall methodology}.

\subsection{Data}
The study uses Call Detail Records of nearly 13 million SIMs from a mobile operator in Sri Lanka. Data was provided for this research by LIRNEasia - a regional ICT policy and regulation think-tank. Data is completely pseudonymized by the operator, where phone numbers are present.

Study also use voluntarily collected CDR and GPS data in a pseudonymized manner from an unbiased sample of mobile users. Data was collected through a mobile application developed for Android devices and users could download the mobile application from the Google play store. Data was collected from more than 700 users for a period of three months from 21-03-2018 to 21-06-2019. This app collects socioeconomic data, CDR data, signal strength data as well as travel data. For validation purposes, the App also collects signal strength and GPS data every 10 minutes and it identifies user's travelling movements automatically through GPS data and prompts the user to indicate their travel purpose.

The main data source used for the purpose of validation is obtained from transportation-related data collected from household surveys within the Western Province of Sri Lanka, which was conducted as part of the CoMTrans study~\cite{comtrans2014}. The Household Visit Survey (HVS) data provide transportation information, including social-demographic records, travel time, trip purposes, travel modes, etc. which can be used for validating the results from CDR trip analysis.

\subsection{Preprocessing}
Preprocessing is done using two filtering techniques. First filter is used for the removal of atypical users. As an example users with low caller activity levels, users with low appearance within the study area (e.g. tourist, visitors). Most of the previous researchers on this was done based on call activity count (CAC). It is required to remove the infrequently visited users who are entering the focused research area, but the removal should be done based on their spatial behaviour but not considering the caller activity counts. 
Our main approach is to use a filter based on spatial behavior, to filter the users. Shannon Entropy is used for the analysis of user's spatial activities.

\begin{equation}
    H=-\sum_{i=1}^{N} P_i\,log_2\,P_i
    \label{Eq:shannon_entropy_equation}
\end{equation}

In the Shannon entropy equation, P refers to the probability that an activity was observed at location i from the set of N locations that the user visits. Number of locations appeared directly affect the behavioral measure. When the total number of locations appeared increase the appearance probability of locations increase by which the entropy value increase. The spatial behavior of 80\% of the users are more of a homogeneous behavior compared to others.

Then the Western Province users were filtered since the study area is the Western Province of Sri Lanka.

\subsection{User Profiling}
One of the major drawbacks of previous research studies is that they consider all users as a single group and use the same model parameters for all users. Especially, when considering mobility patterns at individual level, initially it is essential to profile users based on their socio-economic parameters. Since the travelling behaviour of people highly depend on their socio-economic parameters such as occupation, age, income and etc. As an example, for a public sector worker the general working hours are between 8.30 AM to 4.30 PM and do not work during weekends. But an employee in the private sector may work beyond 4.30 PM and also can appear in work locations during weekends. The behavior of a shift worker or a school worker might be completely different to this. A school employee generally work from 7.30 AM - 1.30 PM but a shift worker may be employed in the night. Hence the behavior can be completely different to each other. Considering the age, a student's travelling pattern cannot be expected from a retired adult and income also has a significant influence on the travelling patterns.

Study selected top features for user profiling based on correlation analysis. The identified features are as follows.
\begin{itemize}
\item Call frequency for each hour
\item Distance for each hour
\item Appeared cell frequency
\item Appearance in weekday or weekends
\end{itemize}
% \hfill

The study uses several classification methods such as Nearest Neighbor, Random Forest, Neural Networks and Support Vector Machine for the data collected from mobile app. Since the dataset has some class imbalance problem, study used SVM weighted classes method (\svm). Then the same models were used to classify the main CDR dataset, as the next step. 

\subsection{Load Sharing Record Identification Interface}
Main aim of this section is the identification of Load sharing records. Initial attempt was to use a speed based filtering method. In speed based filtering a small trajectory data within the corresponding time are inspected and any points which have irrational travel distance or travel speed are labeled as load shared record CDRs. Then the speed is calculated based on two consecutive CDRs for each user and a speed filter was added. This methodology tries to identify the jumps between location sequence.

The speed limit was selected as 120 kmph based on previous studies. But the results were not as expected. In this method precision is good but recall is very low. In other words, the majority of the records which were categorized as load shared record are actually load shared records. But that method missed a lot of load shared records.

Then the study attempts on a new approach to identify load shared records. Main drawback of the predefined speed based filtering method is that speed of the vehicle in the road is not always equal. For an example in the morning 7am-9am average vehicle speed is low compared to 8.30pm-10pm. Generally the speed in one of the main corridors to Colombo at peak hours is 6 kmph but during off peak hours the speed can be high up to 100 kmph. Vehicle speed also depends on the geographical area. In urban cities the vehicle speed is slow compared to remote areas. Our approach is to address both scenarios. Speed limit data was taken from previous transportation origin-destination surveys.
Since the speed limit is based on the time, time frames were defined for our study as, 
07:00-09:00, 09:00-12:00, 12:00-13:00,13:00-16:30,16:30-19:00,19:00-22:00 and 22:00-07:00.
Then the study area is divided based on the Divisional Secretariat Divisions (DSD)and the average speeds were taken based on separate time windows. CDR were filtered based on the $\theta$ values, which represent in kmph, $\theta$ = [0, 5, 10, \dots, 200]
Then the records were evaluated and labeled as Load shared records using GPS data which represent actual movements. Based on this result, the same process was carried out for the main dataset.

\subsection{Reduce Localization Error}
Main part of this research is the identification of the user's location.
CDR data was clustered based on their spatial behaviour to identify different regions of stay for the user (Stay location clusters). The study use DBSCAN to perform the clustering.

Majority of the previous research considered call frequency related data only. However, that does not provide acceptable results. Therefore, this study propose a new approach to minimize user's localization error. It is noted that the assignment of a cell tower to a user depends, among others,  on user's location, signal strength and the call traffic load of the particular cell tower. Following three factors were consider to assign locations for users.
\begin{itemize}
\item Load sharing records
\item Signal strength of the particular cell tower
\item Number of days appeared in the particular cell tower.
\end{itemize}
% \hfill

\begin{figure}
    \centering
    \begin{minipage}{0.45\textwidth}
        \centering
        \includegraphics[width=0.8\textwidth]{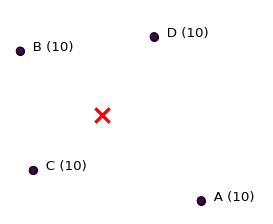} % first figure itself
        \caption{Use locations based on the number of appearances days}
        \label{fig:appearances_days_centroid}
    \end{minipage}\hfill
    \begin{minipage}{0.45\textwidth}
        \centering
        \includegraphics[width=0.8\textwidth]{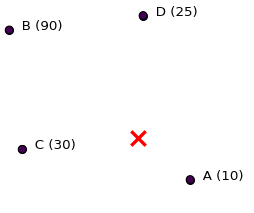} % second figure itself
        \caption{Use locations based on the signal strength}	
	    \label{fig:signal_strength_centroid}
    \end{minipage}
\end{figure}

Consider the following scenario. Assume that User X is connected with cell towers A, B, C, D with an equal number of caller activities. When the weights are based on the caller activities, then the user's location should be the centroid of the considered 4 locations within the particular time period (\figurename~\ref{fig:appearances_days_centroid}).

% \begin{figure}[!hb]	
% 	\centering
% \includegraphics[width=0.2\textwidth]{Images/appearances_days_centroid.png}
% 	\caption{Use locations based on the number of appearances days}	
% 	\label{fig:appearances_days_centroid}
% \end{figure}

In order for the above to happen the signal strength of the cells should be equal to each other. But when the signal strength data of cell towers is taken as the weight, the position of the centroid will change from that of the position taken from caller activity levels as shown in the ~\figurename~\ref{fig:signal_strength_centroid}.

% \begin{figure}[!hb]	
% 	\centering
% \includegraphics[width=0.2\textwidth]{Images/signal_strength_centroid.png}
% 	\caption{Use locations based on the signal strength}	
% 	\label{fig:signal_strength_centroid}
% \end{figure}

Similarly, based on Load shared record frequency, it is possible to assign a location to the particular user.

The study collects signal transmit power for each cell in the study area, and it is used as a measurement to rank the power of connecting to users. $P_{i}$ demonstrate the signal transmitting power of the i\textsuperscript{th} cell.

% The study collects signal strength data every 10 minutes with GPS coordinate using the mobile app. The average of the approximated transmission signal strength is calculated by using Equation~\ref{Eq:friis_transmission_equation}.

% \begin{equation}
% {P_{t}}= {\lambda{P_{r}}}{d^{2}}
% \label{Eq:friis_transmission_equation}
% \end{equation}

Based on those factors, the weight is given by Equation~\ref{Eq:weighted_equation}. For a given cell $i$, Load sharing records, signal transmit power and Number of appearances days are denoted by ${L_{i}}$, ${P_{i}}$, ${C_{i}}$, respectively for particular user.

\begin{equation}
{W_{i}} = {\alpha{L_{i}} + \beta{\frac{1}{P_{i}}} + \gamma{C_{i}}}
\label{Eq:weighted_equation}
\end{equation}

Then, the ${L_{i}}$, ${P_{i}}$ and ${C_{i}}$ values were scaled as in Equation~\ref{Eq:scaling_LSC}.

\begin{equation}
0 \leq L_{i}, ~\frac{1}{P_{i}}, ~C_{i} \leq 1
\label{Eq:scaling_LSC}
\end{equation}

$\alpha$, $\beta$, $\gamma$ are distinct weights calculated for each user segmentation separately within the range mentioned in the Equation~\ref{Eq:scaling_S}.

\begin{equation}
0 \leq \alpha,~ \beta,~ \gamma \leq 1
\label{Eq:scaling_S}
\end{equation}

The study used weighted k-means++ algorithm to find the centroid of the cluster using the weights given by Equation~\ref{Eq:weighted_equation}. Centroids are assigned as user location for a given time.

\begin{table}[!htb]
	\centering
	\caption{User profiling results comparison (P=Precision, R=Recall, F1=F1 Score)}
	\label{tab:ResCompUP}
    \begin{tabular}{|c|c|c|c|c|c|c|c|c|c|c|c|c|c|c|c|c|c|c|}
        \hline
        \multirow{3}{*}{Model} & \multicolumn{3}{c|}{\multirow{2}{*}{\begin{tabular}[c]{@{}c@{}}Full-Time\\ Employees\end{tabular}}} & \multicolumn{3}{c|}{\multirow{2}{*}{\begin{tabular}[c]{@{}c@{}}Part-Time\\ Employees\end{tabular}}} & \multicolumn{3}{c|}{\multirow{2}{*}{Student}} & \multicolumn{3}{c|}{\multirow{2}{*}{Housewife}} & \multicolumn{3}{c|}{\multirow{2}{*}{Retired}} & \multicolumn{3}{c|}{\multirow{2}{*}{Others}} \\
                       & \multicolumn{3}{c|}{} & \multicolumn{3}{c|}{} & \multicolumn{3}{c|}{} & \multicolumn{3}{c|}{} & \multicolumn{3}{c|}{} & \multicolumn{3}{c|}{} \\ \cline{2-19} 
        & P & R & F1 & P & R & F1 & P & R & F1 & P & R & F1 & P & R & F1 & P & R & F1 \\
        \hline
		\ranf & 0.64 & 0.53 & 0.58 & 0.45 & 0.39 & 0.42 & 0.70 & 0.91 & 0.79 & 0.68 & 0.72 & 0.70 & 0.32 & 0.21 & 0.26 & 0.52 & 0.64 & 0.57\\
		\hline
		\ann & 0.70 & 0.44 & 0.54 & 0.55 & 0.40 & \textbf{0.46} & 0.71 & 0.93 & 0.81 & 0.72 & 0.78 & 0.75 & 0.36 & 0.24 & 0.28 & 0.56 & 0.36 & 0.44\\
        \hline
		\svm & 0.71 & 0.66 & \textbf{0.68} & 0.50 & 0.43 & \textbf{0.46} & 0.77 & 0.95 & \textbf{0.85} & 0.71 & 0.81 & \textbf{0.76} & 0.31 & 0.29 & \textbf{0.30} & 0.65 & 0.53 & \textbf{0.58}\\
        \hline
    \end{tabular}
\end{table}

% % results table
% \begin{table*}[!htb]
% 	\centering
% 	\caption{User Profiling Results comparison (P=Precision, R=Recall, F1=F1 Score)}
% 	\label{tab:ResCompUP}
% 	\begin{tabular}{|l|c|c|c|c|c|c|c|c|c|c|c|c|c|c|c|c|c|c|}
% 		\hline
% 		 \multirow{2}{*}{Model} & \multicolumn{3}{c}{Full Time Workers} \vline & \multicolumn{3}{c}{Part Time Workers} \vline & \multicolumn{3}{c}{Student} \vline & \multicolumn{3}{c}{Housewife} \vline & \multicolumn{3}{c}{Retired} \vline & \multicolumn{3}{c}{Others} \vline  \\ 
% 		\hhline{~------------------}
%         & P & R & F1 & P & R & F1 & P & R & F1 & P & R & F1 & P & R & F1 & P & R & F1 \\
%         \hline
% 		\ranf & 0.64 & 0.53 & 0.58 & 0.45 & 0.39 & 0.42 & 0.70 & 0.91 & 0.79 & 0.68 & 0.72 & 0.70 & 0.32 & 0.21 & 0.26 & 0.52 & 0.64 & 0.57\\
% 		\hline
% 		\ann & 0.70 & 0.44 & 0.54 & 0.55 & 0.40 & \textbf{0.46} & 0.71 & 0.93 & 0.81 & 0.72 & 0.78 & 0.75 & 0.36 & 0.24 & 0.28 & 0.56 & 0.36 & 0.44\\
%         \hline
% 		\svm & 0.71 & 0.66 & \textbf{0.68} & 0.50 & 0.43 & \textbf{0.46} & 0.77 & 0.95 & \textbf{0.85} & 0.71 & 0.81 & \textbf{0.76} & 0.31 & 0.29 & \textbf{0.30} & 0.65 & 0.53 & \textbf{0.58}\\
%         \hline
% 	\end{tabular}	
% \end{table*} 
% %end results table          Random Forest, ANN, SVM weighted classes

\section{Results}
\label{Results}

Results of the User Profiling models are shown in Table~\ref{tab:ResCompUP}. The \svm~model showed higher accuracy values for user profiling.

Then Load Sharing Record identification is evaluated. If the last recorded cell tower changes without an actual movement of 100m of the user, study considered that record as a load shared one. The result is shown in Table~\ref{tab:ResCompLSR}. Our approach shows significant accuracy improvement, especially in recall.

To evaluate the Localization result, two approaches were used. First, result were evaluated using mobile app data. To calculate home location error, initially the user's home location which is provided by the user was taken and compared with the location predicted by our method. \figurename~\ref{fig:HomeLocalizationError} indicates the generated results. The same process was carried out for the work location identification of users and the generated results are shown in ~\figurename~\ref{fig:WorkLocalizationError}. Our proposed methodology outperforms existing results in majority of the population. In the study work location was more accurately identify than the home location. The study reveals that 70\% (468) of the users are working in an urban area, but their home locations are in suburban areas. In urban areas cell tower density is high and, in the suburban area cell tower density is low. Therefore, work location error is low compared to the home location.

As the second approach to evaluate the Localization result, Origin-Destination (O-D) matrices~\cite{zhang2010daily} were created from the large (13 million subscribers) CDR dataset. Accordingly, the weekends and holidays were removed from the large CDR dataset. Home activities are defined between 8PM and 5AM~\cite{kung2014exploring} and work activities from 10am to 12pm and 1pm to 4pm~\cite{jiang2013review}. Then the results were compared with HVS O-D matrix data.

Table~\ref{tab:ResCompHVS} indicates the O-D matrices derived from HVS data at district level. This study has been done by covering the Western Province which consists of three districts namely Colombo, Gampaha, Kalutara. As an example cell 1 says that 44\% of users have both their home and work location at the Colombo district, cell 2 says that 3\% of users have their work location within Gampaha and home location within the Colombo district.

Further Table~\ref{tab:ResCompCAR} indicates the user localization based on the number of days the cell tower was connected~\cite{isaacman2011identifying} derived from CDR dataset, which contain nearly 13 million users home-work location distribution. And this table also has the same interpretation as in Table~\ref{tab:ResCompHVS}.

Table~\ref{tab:ResCompOurMethod} indicates the user localization based on our proposed methodology, which considers the user's Load shared record, Signal strength and Number of appearance days and also has the same interpretation as in Table~\ref{tab:ResCompHVS}.

In order to statistically compare the extracted O-D matrices with ground truth (HVS data), study used the Pearson's chi-squared test given by Equation~\ref{Eq:chi_squared}.

\begin{equation}
\tilde{\chi}^2=\sum_{i=1}^{n} \frac{(O_i - E_i)^2}{E_i}
\label{Eq:chi_squared}
\end{equation}

$\chi$ refers to the Pearson's cumulative test statistic, where the 
${E_{i}}$ denotes the Expected value,
${O_{i}}$ the Observation value and 
n is the number of cells in the table. Based on Chi-squared test values, p value for Table~\ref{tab:ResCompCAR} is 0.64 and for Table~\ref{tab:ResCompOurMethod} is 0.88. Our approach outperforms the result, and it shows similar outcomes based on the HVS dataset according to the statistic.

% results table
\begin{table}[!htb]
	\centering
	\caption{Load sharing record identification results comparison}\label{tab:ResCompLSR}
    \begin{tabular}{|l|c|c|c|}
		\hline
        Method & Precision & Recall & F1 \\
		\hline
		Pre-define speed base filter & 0.914 & 0.166 & 0.281 \\
		\hline
		Our Method & 0.864 & 0.728 & \textbf{0.790} \\
        \hline
	\end{tabular}
\end{table} 
%end results table

\begin{figure}
    \centering
    \begin{minipage}{0.45\textwidth}
        \begin{tikzpicture}[scale=0.7]
        \begin{axis}[
            xlabel={Error in Meters},
            ylabel={Percentage of Population},
            x label style={at={(axis description cs:0.5,0.2)}},
            y label style={at={(axis description cs:0.045,0.5)}},
            legend style={at={(0.98,0.25)}, anchor=east},
            xmin=0, xmax=9800,
            ymin=0, ymax=100,
            xmajorgrids=true,
            ymajorgrids=true,
            grid style=dashed,]
        
        \addplot[smooth, color=blue, mark=*,]
            coordinates {(0,0)(3651,70)(6439,80)(9647,90)};
            
        \addplot[smooth, color=red, mark=*,]
            coordinates {(0,0)(1865,70)(3231,80)(4673,90)};
        
        \legend{Call-days based Method, Our Approach}
        \end{axis}
        \end{tikzpicture}
        \caption{Home localization error}
        \label{fig:HomeLocalizationError}
    
    \end{minipage}\hfill
    \begin{minipage}{0.45\textwidth}
        \begin{tikzpicture}[scale=0.7]
        \begin{axis}[%legend pos=south east,
            xlabel={Error in Meters},
            ylabel={Percentage of Population},
            x label style={at={(axis description cs:0.5,0.2)}},
            y label style={at={(axis description cs:0.045,0.5)}},
            legend style={at={(0.98,0.25)}, anchor=east},
            xmin=0, xmax=6700,
            ymin=0, ymax=100,
            xmajorgrids=true,
            ymajorgrids=true,
            grid style=dashed,]
            
        \addplot[smooth, color=blue, mark=*,]
            coordinates {(0,0)(2164,70)(4172,80)(6428,90)};
            
        \addplot[smooth, color=red, mark=*,]
            coordinates {(0,0)(1399,70)(2834,80)(5267,90)};
            
        \legend{Call-days based Method, Our Approach}
        \end{axis}
        \end{tikzpicture}
        \caption{Work localization error}
        \label{fig:WorkLocalizationError}
    \end{minipage}
\end{figure}

\begin{table}[!htb]
	\centering
	\caption{Home work distribution - HVS data}
	\label{tab:ResCompHVS}
	\begin{tabular}{|c|c|c|c|c|}
    \hline
    \multicolumn{2}{|c|}{\multirow{2}{*}{Home/Work}} & \multicolumn{3}{c|}{Trip Attractors} \\
    \cline{3-5} 
    \multicolumn{2}{|c|}{} & Colombo & Gampaha & Kalutara \\
    \hline
    \multirow{3}{*}{Trip Generators} & Colombo & 44\% & 2\% & 1\% \\
    \cline{2-5} & Gampaha & 10\% & 27\% & 1\% \\
    \cline{2-5} & Kalutara & 4\% & 1\% & 10\% \\
    \hline
    \end{tabular}	
\end{table} 
%end results table

% results table
\begin{table}[!htb]
	\centering
	\caption{Home work distribution - Call days based method}
	\label{tab:ResCompCAR}
	\begin{tabular}{|c|c|c|c|c|}
    \hline
    \multicolumn{2}{|c|}{\multirow{2}{*}{Home/Work}} & \multicolumn{3}{c|}{Trip Attractors} \\
    \cline{3-5} 
    \multicolumn{2}{|c|}{} & Colombo & Gampaha & Kalutara \\
    \hline
    \multirow{3}{*}{Trip Generators} & Colombo & 52\% & 2\% & 2\% \\
    \cline{2-5} & Gampaha & 12\% & 18\% & 1\% \\
    \cline{2-5} & Kalutara & 5\% & 1\% & 7\% \\
    \hline
    \end{tabular}	
\end{table} 
%end results table

% results table
\begin{table}[!htb]
	\centering
	\caption{Home work distribution - Proposed methodology}
	\label{tab:ResCompOurMethod}
	\begin{tabular}{|c|c|c|c|c|}
    \hline
    \multicolumn{2}{|c|}{\multirow{2}{*}{Home/Work}} & \multicolumn{3}{c|}{Trip Attractors} \\
    \cline{3-5} 
    \multicolumn{2}{|c|}{} & Colombo & Gampaha & Kalutara \\
    \hline
    \multirow{3}{*}{Trip Generators} & Colombo & 44\% & 3\% & 2\% \\
    \cline{2-5} & Gampaha & 8\% & 30\% & 1\% \\
    \cline{2-5} & Kalutara & 1\% & 1\% & 10\% \\
    \hline
    \end{tabular}	
\end{table} 
%end results table

\section{Conclusion}
\label{conclusion}

As mentioned in the previous studies, real time datasets such as mobile phone traces provide rich information to support transportation planning and operation. Meanwhile, some related limitations should also be addressed when using these datasets in mobility analysis. The most prominent outcome of this research is the formulation of a methodology to reduce the error in localizing the user by considering the load sharing effects of the CDR records and transmit power of the cell towers. This has a significant effect on the travel conclusions obtained from CDR data. The study identifies that there are different optimum speeds for different geographical areas in separate time windows. Findings were validated by comparing with the HVS data. Findings had a significant improvement after reducing the localization error. An additional validation was done using voluntarily collected CDR data. Since the voluntarily collected data also contains GPS data, it provides an accurate way to evaluate localization error. With this, the load sharing effect can be identified easily and hence the accuracy of the findings were also compared with the mobile application data.

Future work is possible through the use of Google traffic data and by setting a speed limit based on the corridors and input load sharing filtering techniques, which are computationally efficient and accurate in result acquisition.\\

\textbf{Acknowledgment.} The authors wish to thank the DataSEARCH Centre for supporting the project which was partially funded by the Senate Research Council under the grant number SRC/LT/2018/08 of the University of Moratuwa.


\bibliographystyle{splncs04}
\bibliography{references}

\end{document}